\documentclass[runningheads]{llncs}
\usepackage{graphicx}
\usepackage{amsmath,amssymb} % define this before the line numbering.
\usepackage{color}
\usepackage[table]{xcolor}
\definecolor{lightgray}{gray}{0.9}
\usepackage[width=122mm,left=12mm,paperwidth=146mm,height=193mm,top=12mm,paperheight=217mm]{geometry}
\begin{document}
% \linenumbers
\pagestyle{headings}
\mainmatter
\def\ECCV16SubNumber{1247}  % Insert your submission number here

\title{Fast Object Localization Using a CNN Feature Map Based Multi-Scale Search} % Replace with your title

\titlerunning{Fast Object Localization Based Multi-Scale Search}

\authorrunning{H. Lee et al.}

\author{Hyungtae Lee\inst{1}, Heesung Kwon\inst{1}, Archith J. Bency\inst{2}, \\and William D. Nothwang\inst{1}}
\institute{U.S. Army Research Laboratory, Adelphi, MD, USA \\ \and University of California, Santa Barbara, CA, USA}

\maketitle

\begin{abstract}
Object localization is an important task in computer vision but requires a large amount of computational power due mainly to an exhaustive multiscale search on the input image.  In this paper, we describe a near real-time multiscale search on a deep CNN feature map that does not use  region proposals. The proposed approach effectively exploits local semantic information preserved in the feature map of the outermost convolutional layer.  A multi-scale search is performed on the feature map by processing all the sub-regions of different sizes using separate expert units of fully connected layers. Each expert unit receives as input local semantic features only from the corresponding sub-regions of a specific geometric shape. Therefore, it contains more nearly optimal parameters tailored to the corresponding shape. This multi-scale and multi-aspect ratio scanning strategy can effectively localize a potential object of an arbitrary size. The proposed approach is fast and able to localize objects of interest with a frame rate of 4 fps while providing improved detection performance over the state-of-the art on the PASCAL VOC 12 and MSCOCO data sets. 
\keywords{object localization, object classification, CNN, multi-scale search, PASCAL VOC, Microsoft COCO}
\end{abstract}

\section{Introduction}

Accurately recognizing objects of interest embedded in images is of great interest to many applications in computer vision.  Recent advances in deep convolutional neural networks are able to provide unprecedented recognition performance mainly due to deep nonlinear exploitation of underlying image data structures.  However, unlike classification localizing objects in images require considerably longer computation time due mainly to an exhaustive search on the input image.
%However, object localization takes considerably long run-time, normally tens of seconds for one image, which makes these approaches impractical for the real-time image/video analytics applications.

Krizhevsky et al.~\cite{AKrizhevskyNIPS12} introduced a deep layered structure that generated breakthrough performance in visual object classification tasks.  The structure referred to as ``deep convolutional neural network (DCNN)'' consists of 8 principal layers which are built on first five convolutional layers and subsequent three fully connected layers, and several supplementary layers.  In this structure, the convolutional layers are the ones that can make the network deep while requiring significantly lesser number of learnable parameters when compared to a network with only fully connected layers.  The multiple cascaded convolutional layers effectively capture nonlinear visual features from both local and global perspectives through consecutive applications of local convolutional filters and max pooling.  The application of the local convolutional filters provides superior performance by hierarchically learning the nonlinear structure of objects of interest embedded in images from a large image database, such as ImageNet~\cite{JDengCVPR09}.

\begin{figure}[t]
    \centering
    \includegraphics[trim=10mm 0mm 10mm 0mm,width=0.8\linewidth]{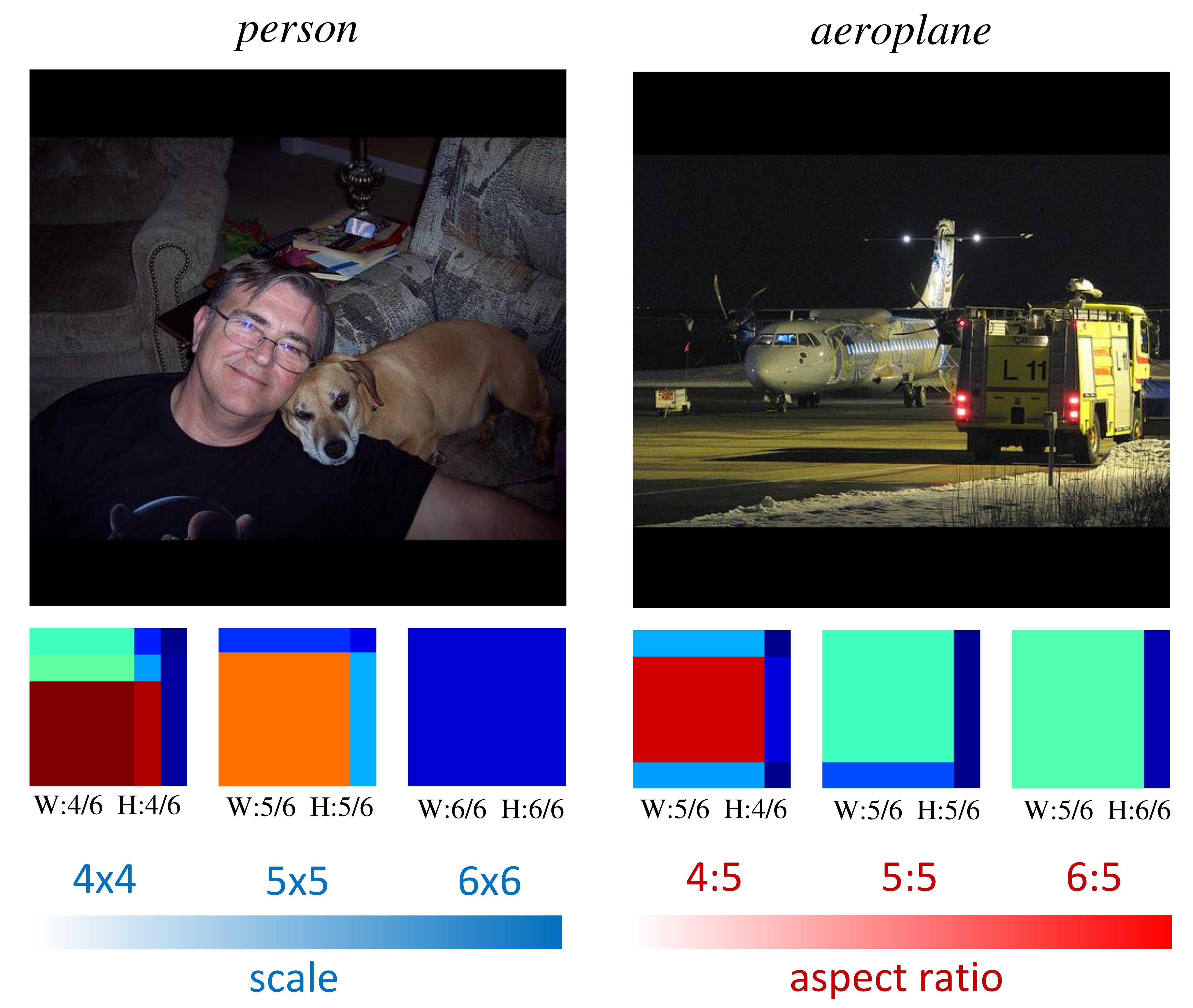}
    \caption{{\bf Effectiveness of the proposed featrue map-based multi-scale and multi-aspect ratio scanning strategy:}  Objects of interest in the images on the left and right sides are {\it person} and {\it aeroplane}, respectively.  On the left side, three classification score maps (red indicates a higher score) from the local windows of three different scales (4$\times$4, 5$\times$5, and 6$\times$6) are generated by using export units of fully connected layers.  Since the 4$\times$4 window on the bottom-left side of the image tightly encloses the person, the classification score of the window on a 4$\times$4 scale has a larger value than other windows of different scales.  On the right side, the local window with the maximum score and an aspect ratio of 4:5 surrounds the aeroplane reflecting the geometrical property of aeroplane.  Thus, the multi-scale and multi-aspect ratio strategy can handle all objects with arbitrary sizes.}
    \label{fig:intro}
\end{figure}

However, object classification by the DCNN is constrained by the fact that the objects in the ImageNet database are roughly located in the center of the image and the object size is relatively large.  This prevents the structure from being directly used for object localization.  One way to use the DCNN for object localization is to use local windows centered on  key points that allow the accurate localizations of objects of interest placed anywhere in the image.  \cite{RGirshickCVPR14,MOquabCVPR14} extract hundreds or thousands of local windows and process each window by rescaling and then applying the DCNN in~\cite{AKrizhevskyNIPS12}.  However, object localization takes considerably long run-time, normally tens of seconds for one image, which makes these approaches impractical for the real-time image/video analytics applications.  %In order to speed up the localization process, several literatures~\cite{MOquabCVPR15,KHePAMI15} are introduced to apply classification layers to multiple local windows in the feature maps of the convolutional layers.  They consider the fact that the feature maps of the convolutional layers contain semantically rich features and coarsely preserve local spatial information of the input image.

In order to reduce the computation time, the proposed approach  processes all the sub-regions (sub-windows) spanning all the locations, sizes, and aspect ratios in the feature map generated by the last convolutional layers.  It performs classification of all the sub-regions  by using separate expert units of fully connected layers, each of which are solely used for the corresponding sub-regions of a particular size and aspect ratio.  Each of the sub-regions is considered a local region with a potential object of interest inside.  Processing the sub-regions in the feature map through the expert units of fully connected layers requires significantly less computational time than repeatedly applying the entire DCNN structure used in \cite{RGirshickCVPR14,MOquabCVPR14}.  As shown in Table~\ref{tab:comp_time}, this multi-scale and multi-aspect ratio window search strategy of independently classifying the sub-regions of different sizes of the feature map makes the proposed method considerably faster than other baselines while providing enhanced accuracy in object localization.

\begin{table}[h]
\begin{center}
\rowcolors{0}{}{lightgray}
\setlength{\tabcolsep}{3pt}
\begin{tabular}{lcc}
\hline
{\small\bf Method} & {\small\bf accuracy (mAP, $\%$)} & {\small\bf time (sec/im)} \\\hline\hline
{\small Oquab15~\cite{MOquabCVPR15}} & {\small 74.5} & 1.3 \\
{\small RCNN~\cite{RGirshickCVPR14}} & {\small 74.8} & {\small 9.0} \\
{\small Fast-RCNN~\cite{RGirshickICCV15}} & {\small 71.3} & {\small 2.1} \\
\hline
{\small Proposed} & {\small 75.4} & {\small 0.23} \\
\hline
\end{tabular}
\end{center}
\caption{Localization accuracy and computation time on PASCAL VOC 2012 validation dataset}
\label{tab:comp_time}
\end{table}

Each of the multiple classification units (mixture of experts) is learned to recognize objects whose size and aspect ratio are similar to those of the corresponding sub-windows.  For instance, 5$\times$4 windows are more appropriate to represent the appearance of the {\it aeroplane} category than 4$\times$5 windows, where the first and second numbers of the dimension indicate its width and height, respectively.  (Please see the example in Figure~\ref{fig:intro}.)   %(Windows of the smallest size (i.e., 3$\times$3) are not considered.)   
We extract the feature maps by applying the convolutional layers of~\cite{AKrizhevskyNIPS12} to a two-level image pyramid which consists of an original image and the double sized image linearly interpolated from the original image.  The size of the feature maps is 6$\times$6 for the original image and 13$\times$13 for the interpolated image.  Therefore, the local windows (4$\times$4 through 6$\times$6) in the 13$\times$13 feature map from the interpolated image are equivalent to the windows of size from 2$\times$2 through 3$\times$3 in the 6x6 feature map of the original input image effectively covering the local window sizes from 2$\times$2 through 6$\times$6.  Consequently, we implement a total of 9 expert units of fully connected layers corresponding to all the windows whose sizes range from 4$\times$4 through 6$\times$6 windows in both the feature maps from the image pyramid.  Figure~\ref{fig:intro} illustrates the effectiveness of this multi-scale and multi-aspect ratio window search strategy for images, in which objects of arbitrary sizes are placed anywhere in the image.

The main contributions of the paper are:
\begin{itemize}
\item We present a novel object detection approach that does not use an exhaustive search or a large number of initial object proposals on the input image.  Instead, a novel multi-scale search on deep CNN feature maps is used resulting in fast object localization with a frame rate 4 fps.
\item Multiple units of fully connected classification layers are introduced for possible detections of different sizes which serve as mixture of expert classifiers, thereby improving detection performance.
%\item The multiple designated units of fully connected classification layers for the corresponding sets of sub-regions of different sizes serving as individual expert classifiers are newly introduced improving overall detection performance.
\end{itemize}

The rest of this paper is organized as follows. Section~\ref{sec:rel_work} presents the related works.  Section~\ref{sec:multi_fc} provides the details of the proposed network.  Experimental results and analysis are presented in Section~\ref{sec:exper} and~\ref{sec:discuss}, respectively.  We conclude the paper in Section~\ref{sec:concl}.

\section{Related work}
\label{sec:rel_work}

\noindent {\bf Literature on the convolutional neural networks: }Since LeCun et al.~\cite{YLeCunNIPS90} introduced convolutional neural netoworks (CNN) in 1990, CNN has been used in various applications in computer vision such as object classification~\cite{AKrizhevskyNIPS12,CSzegedyCVPR15}, object detection~\cite{PSermanetARXIV13,RGirshickCVPR14,MOquabCVPR15,KHePAMI15}, action recognition~\cite{QLeCVPR11,BFernandoCVPR15}, event recognition~\cite{JNgCVPR15,LWangCVPR15,ZWuCVPR15,LYaoCVPR15}, image segementation~\cite{JLongCVPR15,HNohICCV15} and so on.   Convolutional layers have been widely used in deep neural networks because they can make the network deeper without keeping the number of parameters significantly large.  In general, the deeper the network is the better representation it can provide.

Besides the benefit of keeping the number of parameters relatively small, the convolutional layers also provide additional advantages.  Unlike the fully connected layers with fixed input and output dimensions, the convolutional layer allows the structure to be flexible by taking input and output of variable sizes depending on the given tasks.  He et al.~\cite{KHePAMI15} introduced ``spatial pyramid pooling'' which constructs a multi-scale pyramid of feature maps in order to eliminate the requirement that input of CNN is fixed-sized.  Long et al.~\cite{JLongCVPR15} replaced the fully connected layers from~\cite{AKrizhevskyNIPS12} with convolutional layers for semantic segmentation, called a fully convolutional network (FCN).  Oquab et al.~\cite{MOquabCVPR15} also implemented the FCN for object localization.  Moreover, the output of the convolutional layers (i.e., feature maps) preserves local spatial information to a certain degree relative to the original input image.  Figure 6 in Mahendran and Vedaldi~\cite{AMahendranCVPR15} showing reconstructed images from the output of each layer of \cite{AKrizhevskyNIPS12} illustrates the spatial configuration of an input image cannot be recovered after $fc6$ layer.  This finding supports our argument that exploiting the sub-windows of the feature map from the $pool5$ layer along with expert units of fully connected layers is highly efficient for object localization.

\noindent {\bf Literature on using the convolutional neural networks for an object localization: }DCNN in \cite{AKrizhevskyNIPS12} provides high object classification accuracy but is constrained such that relatively large objects of interest are located in the center of the images from the large-scale image database, such as ImageNet.  A large number of local convolutional filters in the multiple convolutional layers learned over millions of training images have an ability to capture a variety of different local appearances caused by different view points and object poses.  However, the convolutional layers may not be effective for the images in which objects are not centrally located.

Several approaches are introduced to address the above issue and apply the DCNN for the object detection problem.  Oquab et al.~\cite{MOquabCVPR14} used a scanning window strategy and apply DCNN to each window in order to localize the object.  \cite{MOquabCVPR15} adapts the last fully connected layer to handle a number of local scanning windows to achieve the localization of objects of interest.  Girshick et al.~\cite{RGirshickCVPR14} apply DCNN to 2000 windows with distinctive objectness characteristics for every test image, which is refered as to ``RCNN''.  However, repeated applications of DCNN greatly increase computational complexity.  Selective search to extract object-like windows in the image used in RCNN also requires about two seconds per an image.  In contrast to the above two approaches, the proposed DCNN is much faster because the convolutional stage is applied only once for the entire image instead of repeatedly applying it for each local scanning window.

\section{Convolutional neural network with multiple units of fully connected layers}
\label{sec:multi_fc}

\begin{figure*}[t]
    \centering
    \includegraphics[trim=10mm 5mm 10mm 10mm,width=\textwidth]{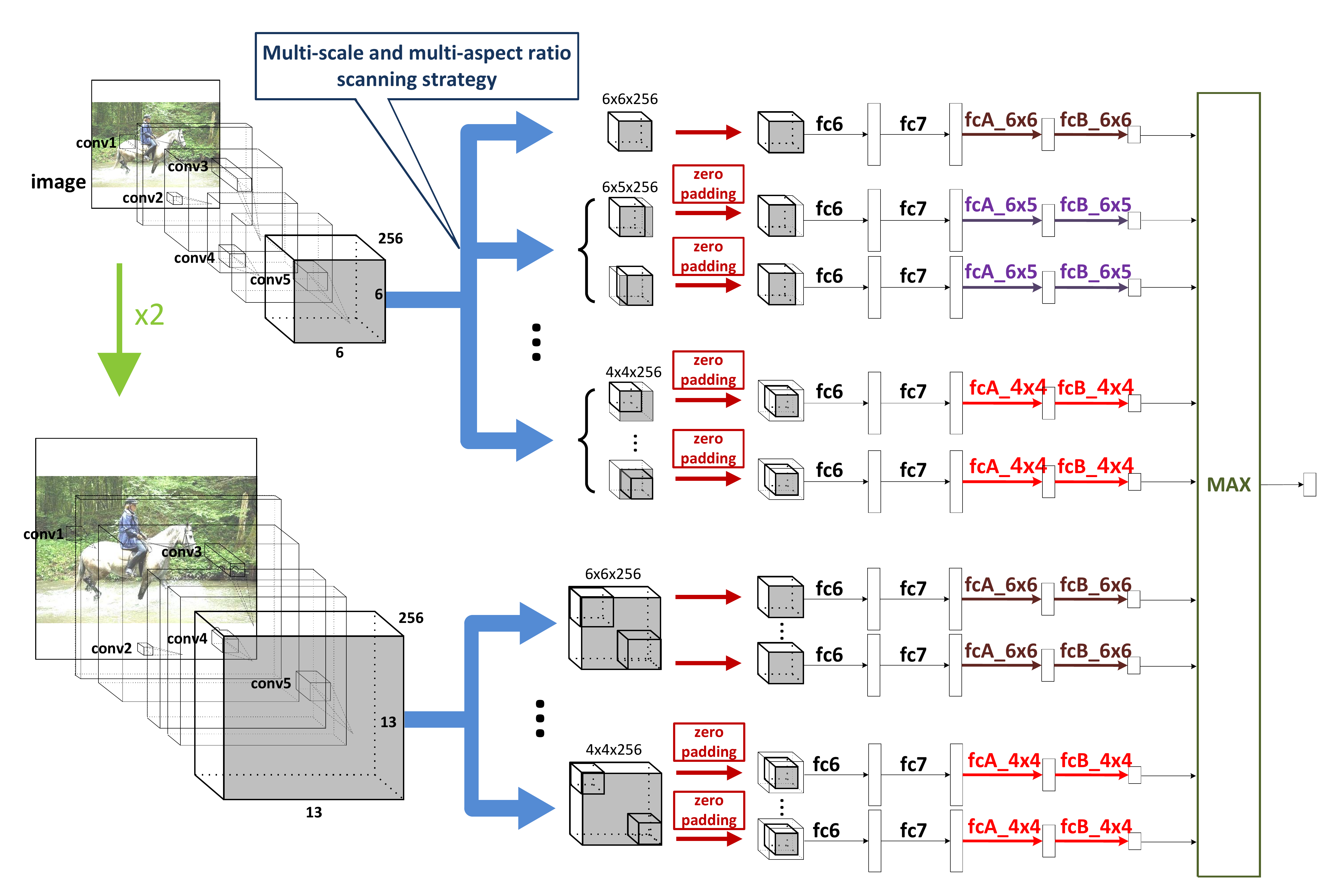}
    \caption{{\bf A block diagram of the proposed DCNN with a two-level image pyramid and the multiple expert units of fully connected layers:} conv1, conv2, conv3, conv4, conv5, fc6, and fc7 are from the architecture of \cite{AKrizhevskyNIPS12} while fcA and fcB are learned.  The proposed scanning strategy  effectively searches sub-windows of different scales and aspect ratios to detect a wide range of objects of different sizes and shape.}
    \label{fig:network}
\end{figure*}

\subsection{Architecture}

The proposed network is built on the architecture of \cite{MOquabCVPR14} that consists of five convolutional layers and four fully connected layers.  The input of the proposed network is a multi-scale image pyramid, as shown in Figure~\ref{fig:network}. The image pyramid is used to effectively handle small objects.  We transfer weights of the first seven layers from DCNN~\cite{AKrizhevskyNIPS12} and fine-tune the last two layers to adapt the network to a small-size target domain dataset.  We denote the convolutional and fully connected layers of the architecture of \cite{MOquabCVPR14} by $conv1,~\cdots,~conv5$, $fc6,~fc7,~fcA,$ and $fcB$, in order.  Since objects of interest can be located anywhere in the target domain images, we intend to exploit coarse spatial correlation between the original input image and the feature map generated by the $conv5$ and the subsequent max pooling stage.  The feature map of each input image is divided into all the possible sub-windows between 4$\times$4 and 6$\times$6, as shown in Figure~\ref{fig:network}, each of which is considered as a candidate region with potential objects of interest inside.  We use multiple independent expert units of fully connected layers, each of which receives the convolutional features of the corresponding sub-window of the feature map separately as input.  Supplementary layers such as ReLU (Rectified Linear Unit), max pooling, local response normalization, dropout, and softmax are selectively applied at the end of or after each layer.

We apply a multi-scale and multi-aspect ratio scanning strategy to the feature maps generated by the convolutional layers.  An inherent characteristic of the convolutional layer is that the local spatial information relative to the original input image is preserved to a certain degree.  To utilize the semantically rich features for representing the input image, we scan sub-windows from a feature map of the last convolutional layer.  The number of sub-windows searched by the scanning strategy, directly related with its computation time, is decided according to the dimension of the feature map.  The scanning strategy searches sub-windows of each feature map whose dimension varies from 4$\times$4 to 6$\times$6.  Sub-windows whose width or height is less than four are not considered due to insufficient feature information. Sub-windows with a width or height over six are not considered because subsequently a fully-connected classification stage receives a 6$\times$6$\times$256 dimensional feature (256 is the number of the filter used in the last convolutional layers).

For each sub-window considered by the scanning strategy, we create a 6$\times$6$\times$256 blob by inserting features in the sub-window into the center of the blob and padding zeros outside the features.  Then, a particular unit of fully connected layers corresponding to the size of the sub-window is applied to the blob and the class scores for objects of interest are calculated.  Scores for all possible sub-windows are collected and a maximum value over the scores for each object category is calculated.  The structure of the proposed network is illustrated in Figure~\ref{fig:network}.

We use a multi-level image pyramid as input to capture small objects in the image, which the unit of the fully connected layers corresponding to smallest sub-window (i.e. 4$\times$4 from the feature map of the original input image)  can not detect.  The original image is rescaled to have the largest side of 227 and then is made to be a square by padding zeros outside of the image.  The aspect ratio of the input image should not be changed since the proposed network is learned as the inherent aspect ratio of objects is preserved.  A higher level image in the pyramid is calculated by resizing the image to twice the width and height (using a linear interpolation), which for instance, indicates a 6$\times$6 sub-window in the higher level image can cover the same region that a 3$\times$3 sub-window in the lower level image can capture. Therefore, a two-level image pyramid consists of two images, one of which has a dimension of 227$\times$227 and the other has a dimension of 454$\times$454.  Figure~\ref{fig:network} illustrates the proposed structure with the two-level image pyramid but can be extended further to accomodate an image pyramid with more than two levels at the expense of computation time.

\subsection{Network training}

\begin{figure}[t]
    \centering
    \includegraphics[trim=10mm 10mm 10mm 10mm,width=0.8\linewidth]{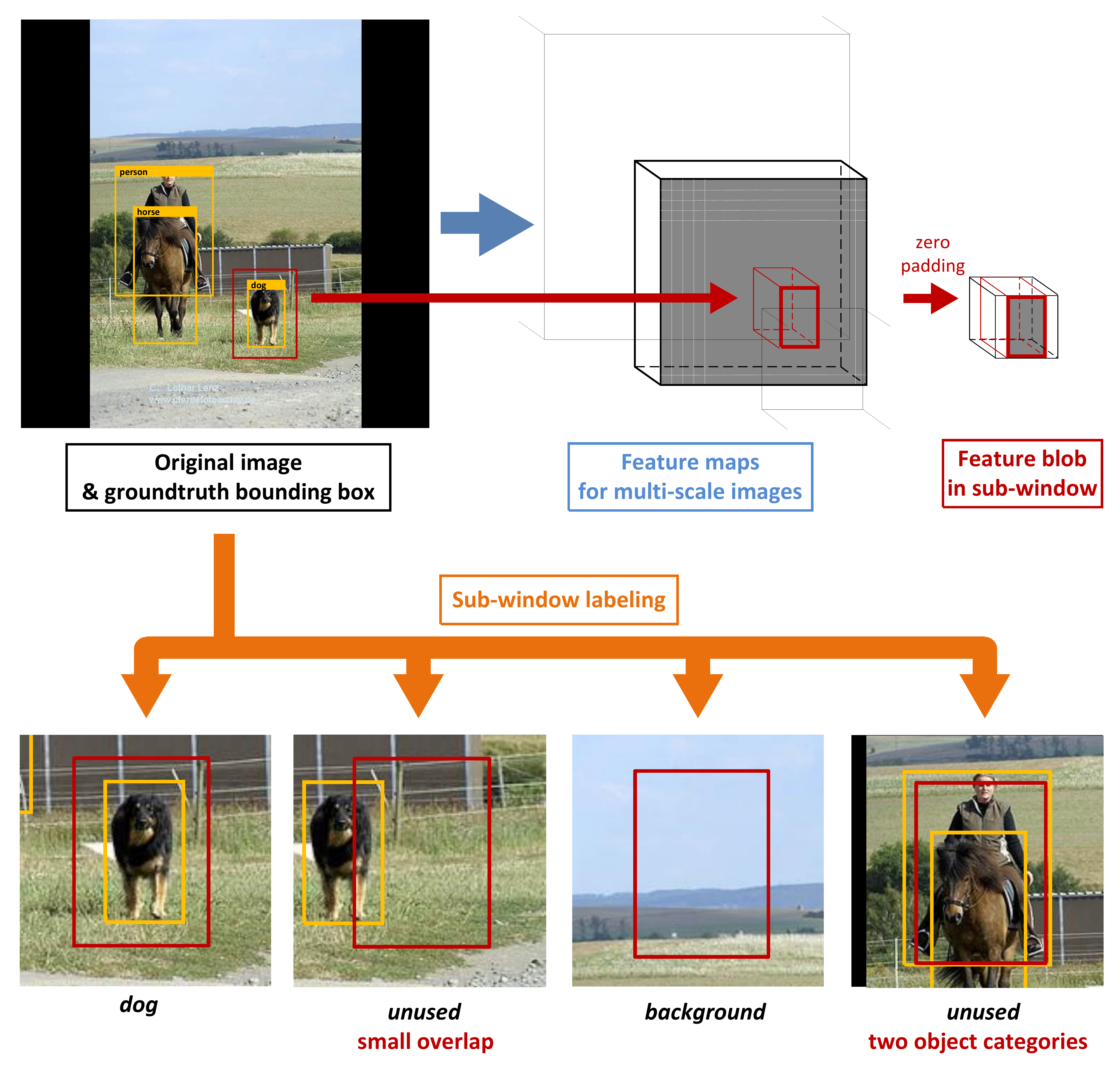}
    \caption{{\bf Collecting sub-windows for training:} Feature maps are generated from a multi-level image pyramid.  (See the blue arrow)  For training a particular unit of fully connected layers, sub-windows with a size corresponding to the unit are collected and then labeled as one of {\it object categories}, {\it background}, or {\it unused}.  This is done by comparing the bounding box corresponding to the sub-window denoted by a red box and groundtruth bounding box denoted by a yellow box. (See the orange arrow.)  Due to the fixed input size of the unit of fully connected layers, 6$\times$6$\times$256 blob is created and the features in the sub-window is filled in the center of the blob.  (See the red arrow.)}
    \label{fig:labeling}
\end{figure}

As we mentioned in the previous section, weights of $conv1,\cdots,~conv5,~fc6,~fc7$ are tranferred from DCNN trained for the ImageNet classification task and the last two fully connected classification layers of each unit are learned for our task that is to localize objects located anywhere in the image.  Each expert unit of fully connected layers is learned with a separate training set because it should have an ability to detect objects with a particular size.  For example, a 4$\times$6 sub-window can express  ``standing persons'' more properly than a 6$\times$4 sub-window.  In contrast, a ``train'' can be expressed better by a 6$\times$4 sub-window.  Table~\ref{tab:one_vs_multiFC} supports effectiveness of using multiple expert units of fully connected layers. 

To collect positive and negative samples, we apply a multi-scale sliding window strategy to the training images by extracting sub-windows and categorizing them into one of object categories, background, or unused.  This strategy introduced in~\cite{MOquabCVPR14} can increase the number of the training samples, which is effective to avoid over-fitting in training.  The training image is normalized to have its largest dimension to be 227 pixels while maintaining the aspect ratio of the image similar to the rescaling of test images.  We define a set of scale factors $\lambda \in \{1,1.3,1.6,2,2.4,2.8,3.2,3.6,4\}$.  For each scale factor, the feature map is computed by using the convolutional layers to the image rescaled again with the factor.  For training a certain unit of fully connected layers, we collect all possible sub-windows of a particular size corresponding to the unit by scanning all scaled images.

To assign a label to each sub-window, we use the bounding box of the sub-window.  We estimate the bounding box of the sub-window in the image coordinate by using the position of the sub-patch in the feature map.  We measure overlapped area $B_{ov}$ between bounding box of the sub-window $B_r$ and ground truth bounding box $B_{gt}$.  Sub-window is labeled as a ``positive'' for a particular object if $B_{ov}/B_r \ge 0.5$ and $B_{ov}/B_{gt} \ge 0.65$.  Otherwise, sub-windows under the condition of $B_{ov}/B_r \le 0.1$ and $B_{ov}/B_{gt} \le 0.1$ are labeled as a ``background''.  A sub-window labeled as a positive for more than one object or not labeled as a positive or a background is unused for training.  All sub-windows labeled as ``background'' are not used due to the training data  becoming imbalanced.  A sub-windows used as ``background'' in training are randomly chosen with a rate $r$ which is specified according to the dataset.  Extracting hard negative samples for the ``background'' class is left for future work.  In experiments, we use $r$ of 0.1 and 0.02 for PASCAL VOC 12 and Microsoft COCO dataset, respectively.

For each sub-window chosen for training, its feature blob is created by inserting features of the last convolutional layer to the center of the blob and padding zero outside the features.  It is the same process with blobs created to be applied to the fully connected layers.  Since pre-trained network depends on the assumption that the object of the interest is roughly centered in the image, the feature blob is inserted in the center of the training blob as well.  The process for labeling sub-windows and creating training blobs is illustrated in Figure~\ref{fig:labeling}.

\section{Experiments}
\label{sec:exper}

\subsection{Dataset and evaluation protocols}

%\begin{figure}[t]
%    \centering
%    \includegraphics[trim=10mm 10mm 10mm 10mm,width=\linewidth]{dataset.pdf}
%    \caption{Several exmaple images from {\it horse} category in three datasets: ImageNet~\cite{JDengCVPR09}, PASCAL VOC 12~\cite{pascal-voc-2012} , and Microsoft COCO~\cite{TLinCVPR15} datasets.}
%    \label{fig:dataset}
%\end{figure}

The proposed network is evaluated on two tasks which are object classification and localization on PASCAL VOC 12 dataset~\cite{pascal-voc-2012} and Microsoft COCO dataset~\cite{TLinCVPR15}.  Object classification is to test an image if it contains an object of interest and object localization is to search locations of the object in the image.  In the target datasets that objects can be anywhere in images, object classification performance is closely associated with object localization performance.  It is because a high performance detector such as CNN has few false positive detections that incorrectly detect background as an object of interest but, by chance, the object is located in other place in the image.  Compared to ImageNet dataset~\cite{JDengCVPR09}, target datasets contain a relatively small size of images, which is not enough to avoid overfitting in training the deep-layered network.  We should use either PASCAL VOC 12 dataset or Microsoft COCO dataset rather than ImageNet which is not approapriate to evaluate object localization due to its inherent image characteric.  %(See Figure~\ref{fig:dataset} which shows some example images of the {\it horse} category in the ImageNet, PASCAL VOC 12, and Microsoft COCO datasets.)  
Overfitting issue is solved by utilizing fine-tuning as in~\cite{MOquabCVPR14}.  We use Caffe~\cite{jia2014caffe} as the framework where the proposed network is implemented.

PASCAL VOC 12 dataset consisting of approximately 22k images contains 20 object categories and provides {\tt trainval} and {\tt test} for training the network and evaluating test images.  Microsoft COCO dataset contains 80k images for training and 40k for validation and 80 object categories.

\subsection{Object classification}

\begin{table}[htbp]
\begin{center}
\rowcolors{0}{white}{lightgray}
\setlength{\tabcolsep}{3pt}
\begin{tabular}{l|cc|cc}
{\bf {\small Obj Classif.}} & {\small Oquab14~\cite{MOquabCVPR14}} & {\small Oquab15~\cite{MOquabCVPR15}} & {\small MultiFC-2} & {\small MultiFC-3} \\
\hline\hline
{\small aero} & {\small 94.6} & {\small 96.7} & {\small 92.2}  & {\small 93.5} \\
{\small bike} & {\small 82.9} & {\small 88.8} & {\small 78.1} & {\small 81.9} \\
{\small bird} & {\small 88.2} & {\small 92.0} & {\small 83.0} & {\small 86.6} \\
{\small boat} & {\small 84.1} & {\small 87.4} & {\small 77.2}  & {\small 79.0} \\
{\small bottle} & {\small 60.3} & {\small 64.7} & {\small 44.0}  & {\small 57.2} \\
{\small bus} & {\small 89.0} & {\small 91.1} & {\small 84.7}  & {\small 86.8} \\
{\small car} & {\small 84.4} & {\small 87.4} & {\small 74.3}  & {\small 80.8} \\
{\small cat} & {\small 90.7} & {\small 94.4} & {\small 88.5}  & {\small 91.3} \\
{\small chair} & {\small 72.1} & {\small 74.9} & {\small 57.7}  & {\small 62.5} \\
{\small cow} & {\small 86.8} & {\small 89.2} & {\small 67.2}  & {\small 70.8} \\
{\small table} & {\small 69.0} & {\small 76.3} & {\small 66.6}  & {\small 68.3} \\
{\small dog} & {\small 92.1} & {\small 93.7} & {\small 88.4}  & {\small 91.1} \\
{\small horse} & {\small 93.4} & {\small 95.2} & {\small 82.3}  & {\small 83.3} \\
{\small mbike} & {\small 88.6} & {\small 91.1} & {\small 84.9}  & {\small 87.1} \\
{\small person} & {\small 96.1} & {\small 97.6} & {\small 90.8}  & {\small 96.1} \\
{\small plant} & {\small 64.3} & {\small 66.2} & {\small 53.6}  & {\small 62.8} \\
{\small sheep} & {\small 86.6} & {\small 91.2} & {\small 73.7}  & {\small 76.2} \\
{\small sofa} & {\small 69.0} & {\small 70.0} & {\small 53.0}  & {\small 54.2} \\
{\small train} & {\small 91.1} & {\small 94.5} & {\small 86.2}  & {\small 87.6} \\
{\small tv} & {\small 79.8} & {\small 83.7} & {\small 72.7}  & {\small 76.9} \\
\hline
{\small mean} & {\small 82.8} & {\small 86.3} & {\small 75.0} & {\small 78.7} \\
\end{tabular}
\end{center}
\caption{Object classification performance on PASCAL VOC 2012 test dataset~\cite{pascal-voc-2012}}
\label{tab:obj_classifi}
\end{table}

\begin{figure*}[htbp]
    \centering
    \includegraphics[trim=20mm 15mm 20mm 10mm,width=0.98\linewidth]{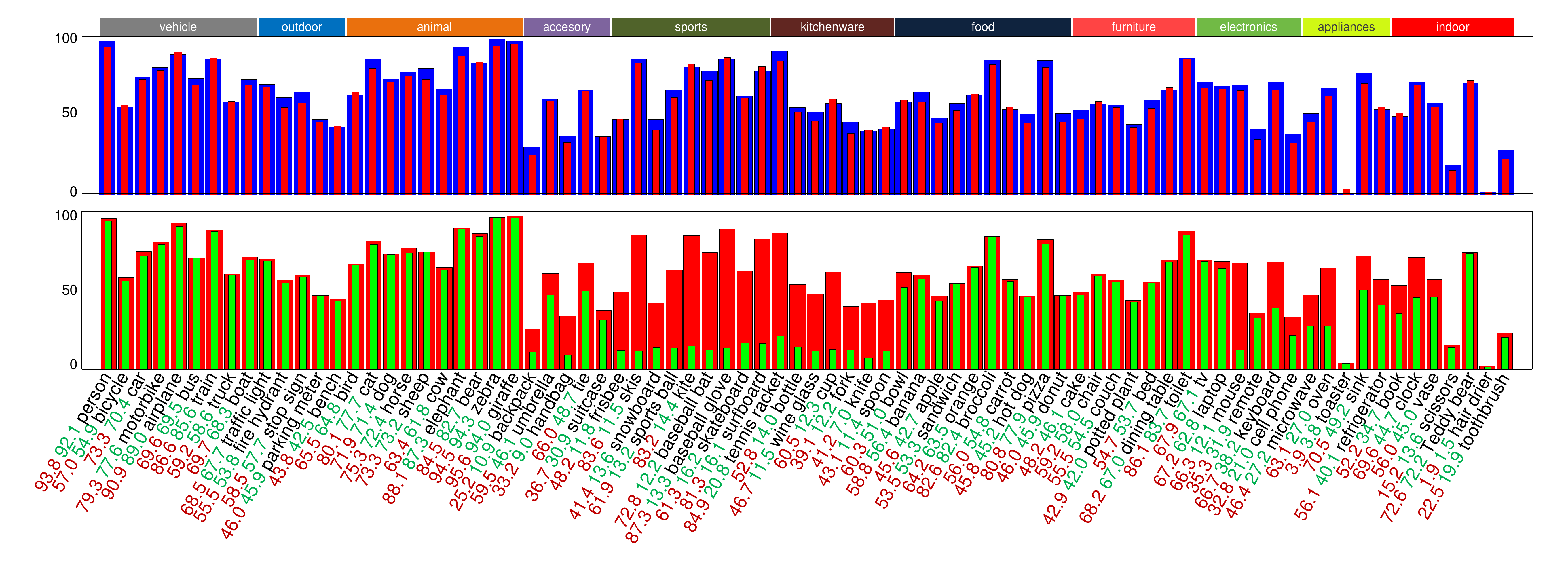}
    \caption{{\bf Object classification and localization performance on Microsoft COCO validation dataset~\cite{TLinCVPR15}:} The plot in the first row compares object classification performance between the proposed network and \cite{MOquabCVPR15} indicated by red and blue bars, respectively.  In the second row, the object localization performance (indicated by green bars) of the proposed network is compared to the object classification performance.  The values in red and green beside each object category along the x axis indicate classification and localization performance (mAP) of the proposed network, respectively.}
    \label{fig:coco_obj_classifi}
\end{figure*}

We apply the proposed network to both target datasets and calculate mean of average precision (mAP) over all object categories.  Table~\ref{tab:obj_classifi} shows the object classification performance of baselines as well as the proposed network on PASCAL VOC 12.  As baselines, we use two CNN-based methods developed by Oquab et al.~\cite{MOquabCVPR14,MOquabCVPR15}.  \cite{MOquabCVPR15} presents the state-of-the art performance in both object classification and localization on the PASCAL VOC 12 dataset.  The plot in the first low in Figure~\ref{fig:coco_obj_classifi} compares object classification performance between the state-of-the art (Oquab et al.~\cite{MOquabCVPR15}) and the proposed network for each object category on Microsoft COCO dataset.

\subsection{Object localization}

\begin{table}[t]
\begin{center}
\rowcolors{0}{white}{lightgray}
\setlength{\tabcolsep}{3pt}
\begin{tabular}{l|ccc|cc}
{\bf {\small Obj local.}} & {\small Oquab15~\cite{MOquabCVPR15}} & {\small RCNN~\cite{RGirshickCVPR14}\footnotemark[1]} & {\small Fast-RCNN~\cite{RGirshickICCV15}\footnotemark[1]} & {\small MultiFC-2} & {\small MultiFC-3} \\
\hline\hline
{\small aero} & {\small 90.3} & {\small 92.0} & {\small 79.2} & {\small 87.7} & {\small 92.9} \\
{\small bike} & {\small 77.4} & {\small 80.8} & {\small 74.7} & {\small 77.3} & {\small 79.7} \\
{\small bird} & {\small 81.4} & {\small 80.8} & {\small 76.2} & {\small 79.8} & {\small 88.7} \\
{\small boat} & {\small 79.2} & {\small 73.0} & {\small 65.8} & {\small 74.4} & {\small 76.5} \\
{\small bottle} & {\small 41.1} & {\small 49.9} & {\small 39.4} & {\small 64.2} & {\small 67.8} \\
{\small bus} & {\small 87.8} & {\small 86.8} & {\small 82.3} & {\small 91.3} & {\small 93.0} \\
{\small car} & {\small 66.4} & {\small 77.7} & {\small 64.8} & {\small 80.1} & {\small 82.2} \\
{\small cat} & {\small 91.0} & {\small 87.6} & {\small 85.7} & {\small 75.7} & {\small 90.7} \\
{\small chair} & {\small 47.3} & {\small 50.4} & {\small 54.5} & {\small 55.7} & {\small 51.3} \\
{\small cow} & {\small 83.7} & {\small 72.1} & {\small 77.2} & {\small 66.7} & {\small 66.6} \\
{\small table} & {\small 55.1} & {\small 57.6} & {\small 58.8} & {\small 65.2} & {\small 64.5} \\
{\small dog} & {\small 88.8} & {\small 82.9} & {\small 85.1} & {\small 83.5} & {\small 87.3} \\
{\small horse} & {\small 93.6} & {\small 79.1} & {\small 86.1} & {\small 78.5} & {\small 78.4} \\
{\small mbike} & {\small 85.2} & {\small 89.8} & {\small 80.5} & {\small 84.6} & {\small 84.1} \\
{\small person} & {\small 87.4} & {\small 88.1} & {\small 76.6} & {\small 89.8} & {\small 95.0} \\
{\small plant} & {\small 43.5} & {\small 56.1} & {\small 46.7} & {\small 61.1} & {\small 62.8} \\
{\small sheep} & {\small 86.2} & {\small 83.5} & {\small 79.5} & {\small 78.1} & {\small 80.9} \\
{\small sofa} & {\small 50.8} & {\small 50.1} & {\small 68.3} & {\small 46.9} & {\small 46.7} \\
{\small train} & {\small 86.8} & {\small 81.5} & {\small 85.0} & {\small 90.1} & {\small 88.3} \\
{\small tv} & {\small 66.5} & {\small 76.6} & {\small 60.0} & {\small 77.8} & {\small 78.1} \\
\hline
{\small mean} & {\small 74.5} & {\small 74.8} & {\small 71.3} & {\small 75.4} & {\bf {\small 77.8}} \\
\end{tabular}
\end{center}
\caption{Object localization performance on PASCAL VOC 2012 validation dataset~\cite{pascal-voc-2012}}
\label{tab:obj_local}
\end{table}

\begin{table}[t]
\begin{center}
\rowcolors{0}{}{lightgray}
\setlength{\tabcolsep}{3pt}
\begin{tabular}{|l|ccccc|}
\hline
& \multicolumn{5}{c|}{{\footnotesize Overlap threshold}} \\ 
\cline{2-6}
{\footnotesize Method} & {\footnotesize 0.1} & {\footnotesize 0.15} & {\footnotesize 0.25} & {\footnotesize 0.3} & {\footnotesize 0.5} \\\hline\hline
{\footnotesize Oquab15~\cite{MOquabCVPR15} + Active Segmentation~\cite{AMishraICCV09}}~& {\footnotesize 17.6} & {\footnotesize 13.6} & {\footnotesize 9.1} & {\footnotesize 7.3} &  {\footnotesize 3.3} \\
{\footnotesize Oquab15~\cite{MOquabCVPR15} + Selective Search~\cite{JUijlingsIJCV13}}~& {\footnotesize 43.5} & - & - & {\footnotesize 27.5} & {\footnotesize 11.7} \\\hline
{\footnotesize MultiFC-2 level} & {\footnotesize 36.1} & {\footnotesize 34.8} & {\footnotesize 30.6} & {\footnotesize 27.7} & {\footnotesize 9.2} \\
{\footnotesize MultiFC-3 level} & {\footnotesize {\bf 49.6}} & {\footnotesize {\bf 48.3}} & {\footnotesize {\bf 43.9}} & {\footnotesize {\bf 40.7}} & {\footnotesize {\bf 15.4}} \\

\hline

\end{tabular}
\end{center}
\caption{Object localization performance with respect to various thresholds based on intersection over union between detection boundingbox and groundtruth bounding box on PASCAL VOC 2012 validation dataset~\cite{pascal-voc-2012}.}
\label{tab:obj_det}
\end{table}

To evaluate object localization, \cite{MOquabCVPR15} introduces a localization criterion that if the location of the highest score in the image falls inside the groundtruth bounding box with extra 18 pixel tolerance to account for the pooling ratio of the network, the image is classified as true positive.  This criterion is useful to evaluate object localization performance for the proposed approach which does not estimate an object bounding box.  Since this criterion can be used to separate correct classifications from false positives, localization performance based on this criterion is likely to be the more accurate classification performance.  We also use the standard object criterion for object localization which is based on the intersection between detection bounding box and groundtruth bounding box.  Since an evaluation server for PASCAL VOC 12 dataset does not calculate the performance based on the first criterion, we divide {\tt trainval} into {\tt train} set for training and {\tt val} set for testing the networks.

\footnotetext[1]{Similar to the evaluation of object localization performance of RCNN in~\cite{MOquabCVPR15}, we use the proposal bounding box with the maximum score per class and an image for evaluation.}

Table~\ref{tab:obj_local} presents the object localization performance of the proposed network and baselines (Oquab et al.~\cite{MOquabCVPR15}, RCNN~\cite{RGirshickCVPR14}, and Fast-RCNN~\cite{RGirshickICCV15}) under the first criterion.  In Table~\ref{tab:obj_det}, we compare the performance of detecting the extent of objects among the proposed network and two baselines under various overlap thresholds.  To produce detection results of \cite{MOquabCVPR15}, several approaches such as active segmentation~\cite{AMishraICCV09} and selective search~\cite{JUijlingsIJCV13} are employed for obtaining object proposals.  For each proposal, classification scores within the proposal bounding box are collected for evaluation.  The proposed network estimates the detection bounding boxes from a sub-window location and its size for each sub-window.  Figure~\ref{fig:qual_eval} shows example images for all the categories of PASCAL VOC 12 as well as corresponding classification score maps.  Table~\ref{tab:obj_local_coco} presents performance of both object classification and localization under the first crietrion on Microsoft COCO dataset.  The plot in the second row in Figure~\ref{fig:coco_obj_classifi} compares object classification and localization performance of the proposed network.

\begin{table}[t]
\begin{center}
\rowcolors{0}{}{lightgray}
\begin{tabular}{|l|c|c|c|}
\hline
& Classification & Localization \\\hline\hline
Oquab15~\cite{MOquabCVPR15} & {\bf 62.8} & 41.2 \\
MultiFC-3 level & 60.4 & {\bf 45.8} \\\hline

\end{tabular}
\end{center}
\caption{Object classification and localization performance on Microsoft COCO dataset~\cite{TLinCVPR15}}
\label{tab:obj_local_coco}
\end{table}

\noindent {\bf Searching the object location using the maximum classification score:} In order to use the first criterion, we compute the classification score across all locations in the image and search the location with the maximum score for a particular object category.  For each pixel in the image, we collect all detections containing that pixel.  Confidence score for the pixel $x$ is computed as

\begin{eqnarray}
sc(x) = \frac{1}{M}\sum_{i~s.t.~x~\in~bbox_i}{sc_i^n} \\\nonumber
x^{*} = \arg \max_{x} sc(x),
\end{eqnarray}
where $M$ is a total number of detections which the location $x$ is in.  $sc(x)$ and $sc_i$ indicate the overall score for position $x$ and the confidence score of $i^{th}$ detection whose bounding box is indicated by $bbox_i$, respectively.  $x^{*}$ is the location with the maximum classification score in the image.  We use five as $n$  in order to suppress the effect of low confident detections.

\section{Discussion}
\label{sec:discuss}

\begin{figure*}[h]
    \centering
    \includegraphics[trim=10mm 10mm 10mm 10mm,width=0.99\textwidth]{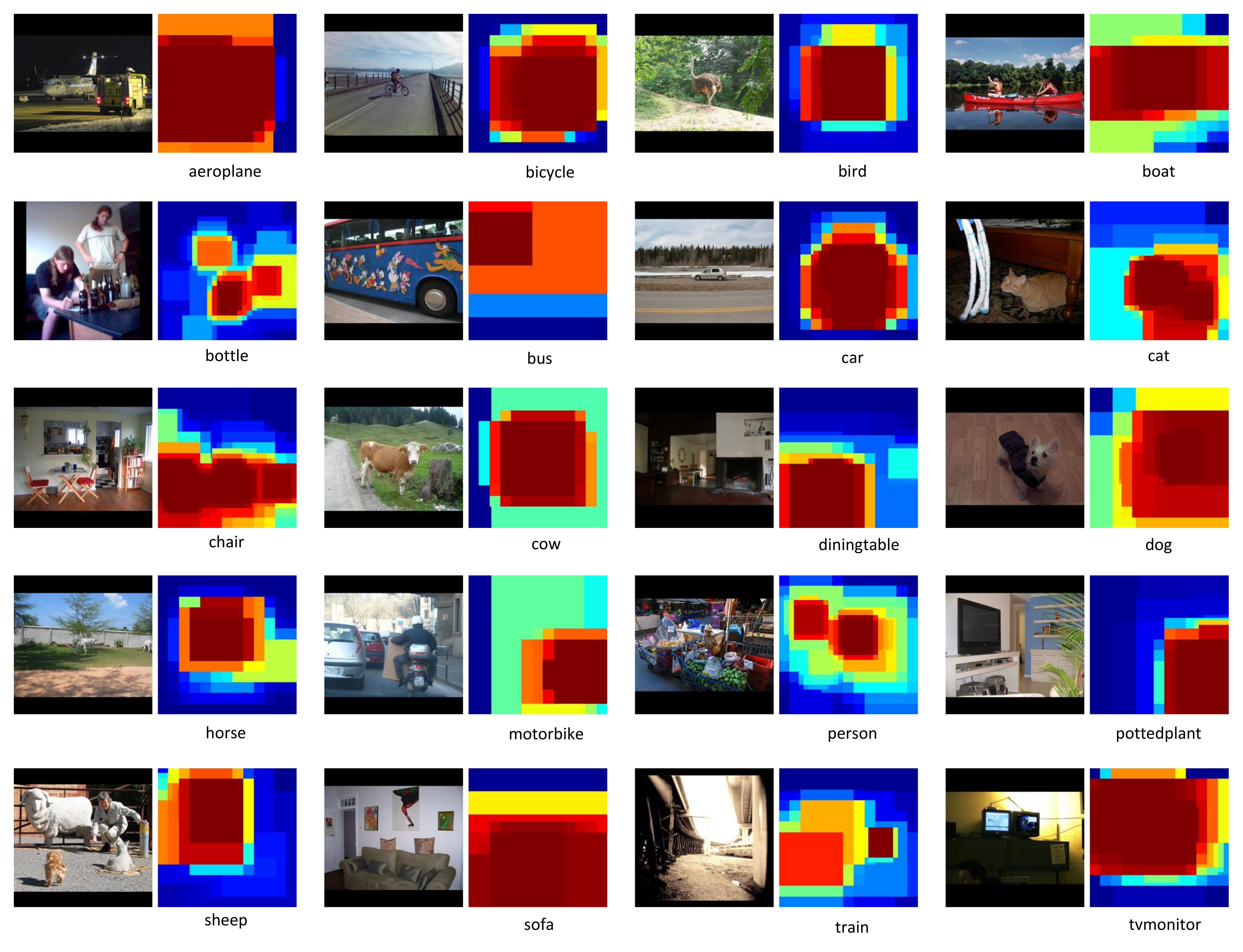}
    \caption{Example images and their corresponding classification score maps (generated by the proposed network) for 20 object categories on PASCAL VOC 12~\cite{pascal-voc-2012}.}
    \label{fig:qual_eval}
\end{figure*}

\noindent {\bf Performance and computation time:} For both datasets, the proposed multi-scale and multi-aspect ratio scanning strategy outperforms all the baselines including RCNN~\cite{RGirshickCVPR14} and fast-RCNN~\cite{RGirshickICCV15} in object localization.  Notably, the object localization performance estimated using the sub-window-based bounding boxes outperforms the approach combining \cite{MOquabCVPR15} with object proposals by the selective search, as shown in Table~\ref{tab:obj_det}.   Figure~\ref{fig:qual_eval} shows that the sub-window with the maximum classification score estimated by the proposed network tends to enclose an object of interest.  As future work, a bounding box regression model can be employed to estimate accurate object bounding box.  However, the proposed network provides slightly lower classification performance than \cite{MOquabCVPR15}.  The small performance drop in classification is primarily caused by using lesser number of sub-windows when compared to the exhaustive scanning.

%For both datasets, our proposed network provides slightly lower classification performance than \cite{MOquabCVPR15}.  The small performance drop in classification is primarily caused by using lesser number of sub-windows when compared to the exhaustive scanning.  

%However, the proposed multi-scale and multi-aspect ratio scanning strategy outperforms all the baselines including RCNN~\cite{RGirshickCVPR14} and fast-RCNN~\cite{RGirshickICCV15} in object localization.  Notably, the object localization performance estimated using the sub-window-based bounding boxes outperforms the approach combining \cite{MOquabCVPR15} with object proposals by the selective search, as shown in Table~\ref{tab:obj_det}.  Figure~\ref{fig:qual_eval} shows that the sub-window with the maximum classification score estimated by the proposed network tends to enclose an object of interest.  As future work, a bounding box regression model can be employed to estimate accurate object bounding box.  

The computation time of the proposed network based on a two-level image pyramid is significantly faster than the baselines as shown in Table~\ref{tab:comput_time}.  The computation time for the proposed network and baselines is measured by using Caffe framework and an NVIDIA GTX TITAN X Desktop GPU.  The proposed network with a three-level image pyramid presents improved accuracy over baselines and a two-level image pyramid (by 2.6 $\%$ for classification and 2.4 $\%$ for localization) but the computation time was slower than one with a two-level image pyramid as expected.

\begin{table}[t]
\begin{center}
\rowcolors{0}{}{lightgray}
\begin{tabular}{lc}
\hline
Method & Comput. time (sec./im) \\\hline\hline
Oquab15~\cite{MOquabCVPR15} & 1.3 \\
RCNN~\cite{RGirshickCVPR14} & 9.0 \\
Fast-RCNN~\cite{RGirshickICCV15} & 2.1 \\\hline
MultiFC-2 level & {\bf 0.23} \\
MultiFC-3 level & 1.58 \\\hline

\end{tabular}
\end{center}
\caption{Computation time of object localization for the proposed network and baselines in test time.}
\label{tab:comput_time}
\end{table}

\begin{table}[t]
\begin{center}
\rowcolors{0}{}{lightgray}
\begin{tabular}{lc}
\hline
Method & Object Localization (mAP) \\\hline\hline
SingleFC & 72.5 \\
MultiFC & {\bf 77.8} \\\hline

\end{tabular}
\end{center}
\caption{The performance of object localization by using single unit of fully connected layers vs. multiple units of fully connected layers (evaluated on PASCAL VOC 12 validation set).}
\label{tab:one_vs_multiFC}
\end{table}

\noindent {\bf Effectiveness of multiple expert units of fully connected layers:} To evaluate the effectiveness of multiple expert units of fully connected layers, we implemented a single unit of fully connected layers which is learned to capture all the appearance of objects with various sizes.  For training the single unit, we collected all training sub-windows used for learning all individual units of fully connected layers.  Table~\ref{tab:one_vs_multiFC} shows that multiple units outperform by 5.3 $\%$ to the single unit in the object localizaion evaluation.  It supports that learning by collecting objects of a particular scale and aspect ratio is effective, which leads to implement the proposed mixture of expert classifiers.

\section{Conclusions}
\label{sec:concl}

This paper presents a fast object localization approach based on the deep convolutional neural network (DCNN) that can provide improved localization performance over the state-of-the art.  The proposed network achieves a frame rate of as fast as 4 fps, which is significantly faster than other CNN-based object localization baselines.  The fast processing time is achieved by using a multi-scale search on deep CNN feature maps instead of relying on an exhaustive search or a large number of initial object proposals on the input image.  The enhanced object localization performance primarily comes from using the multiple expert units of fully connected classification layers that can effectively improve localization of objects in different scales and aspect ratios.

\bibliographystyle{splncs}
\bibliography{egbib}

\begin{thebibliography}{10}

\bibitem{AKrizhevskyNIPS12}
Krizhevsky, A., Sutskever, I., Hinton, G.E.:
\newblock Imagenet classification with deep convolutional neural networks.
\newblock In: NIPS. (2012)

\bibitem{JDengCVPR09}
Deng, J., Socher, R., Li, L.J., Li, K., Fei-Fei, L.:
\newblock Imagenet: A large-scale hierarchical image database.
\newblock In: CVPR. (2009)

\bibitem{RGirshickCVPR14}
Girshick, R., Donahue, J., Darrell, T., Malik, J.:
\newblock Rich feature hierarchies for accurate object detection and semantic
  segmentation.
\newblock In: CVPR. (2014)

\bibitem{MOquabCVPR14}
Oquab, M., Bottou, L., Laptev, I., Sivic, J.:
\newblock Learning and yransferring mid-level image representations using
  convolutional neural networks.
\newblock In: CVPR. (2014)

\bibitem{MOquabCVPR15}
Oquab, M., Bottou, L., Laptev, I., Sivic, J.:
\newblock Is object localization for free? - weakly-supervised learning with
  convolutional neural networks.
\newblock In: CVPR. (2015)

\bibitem{RGirshickICCV15}
Girshick, R.:
\newblock Fast {R-CNN}.
\newblock In: ICCV. (2015)

\bibitem{YLeCunNIPS90}
LeCun, Y., Boser, B., Denker, J., Henderson, D., Howard, R., Hubbard, W.,
  Jackel, L.:
\newblock Handwritten digit recognition with a back-propagation network.
\newblock In: NIPS. (1990)

\bibitem{CSzegedyCVPR15}
Szegedy, C., Liu, W., Jia, Y., Sermanet, P., Reed, S., Anguelov, D., Erhan, D.,
  Vanhoucke, V., Rabinovich, A.:
\newblock Going deeper with convolutions.
\newblock In: CVPR. (2015)

\bibitem{PSermanetARXIV13}
Sermanet, P., Eigen, D., Zhang, X., Mathieu, M., Fergus, R., LeCun, Y.:
\newblock Overfeat: Integrated recognition, localization and detection using
  convolutional networks.
\newblock In: arXiv:1312.6229. (2013)

\bibitem{KHePAMI15}
He, K., Zhang, X., Ren, S., Sun, J.:
\newblock Spatial pyramid pooling in deep convolutional networks for visual
  recognition.
\newblock IEEE Transactions on Pattern Recognition and Machine Intelligence
  (PAMI) \textbf{37} (2015)  1904--1916

\bibitem{QLeCVPR11}
Le, Q.V., Zou, W.Y., Yeung, S.Y., Ng, A.Y.:
\newblock Learning hierarchical invariant spatio-temporal features for action
  recognition with independent subspace analysis.
\newblock In: CVPR. (2011)

\bibitem{BFernandoCVPR15}
Fernando, B., Gavves, E., Orami, J., M., T.T., Ghodrat, A.:
\newblock Modeling video evolution for action recognition.
\newblock In: CVPR. (2015)

\bibitem{JNgCVPR15}
Ng, J.Y.H., Hausknecht, M., Vijayanarasimhan, S., Vinyals, O., Monga, R.,
  Toderici, G.:
\newblock Beyond short snippets: deep networks for video classification.
\newblock In: CVPR. (2015)

\bibitem{LWangCVPR15}
Wang, L., Wang, Z., Du, W., Qiao, Y.:
\newblock Object-scene convolutional neural networks for event recognition in
  images.
\newblock In: CVPR. (2015)

\bibitem{ZWuCVPR15}
Xu, Z., Yang, Y., Hauptmann, A.G.:
\newblock A discriminative {CNN} video representation for event detection.
\newblock In: CVPR. (2015)

\bibitem{LYaoCVPR15}
Yao, L., Torabi, A., Cho, K., Ballas, N., Pal, C., Larochelle, H., Courville,
  A.:
\newblock Describing videos by exploiting temporal structure.
\newblock In: CVPR. (2015)

\bibitem{JLongCVPR15}
Long, J., Shelhamer, E., Darrell, T.:
\newblock Fully convolutional networks for semantic segmentation.
\newblock In: CVPR. (2015)

\bibitem{HNohICCV15}
Noh, H., Hong, S., Han, B.:
\newblock Learning deconvolution network for semantic segmentation.
\newblock In: ICCV. (2015)

\bibitem{AMahendranCVPR15}
Mahendran, A., Vedaldi, A.:
\newblock Understanding deep image representations by inverting them.
\newblock In: CVPR. (2015)

\bibitem{pascal-voc-2012}
Everingham, M., Van~Gool, L., Williams, C.K.I., Winn, J., Zisserman, A.:
\newblock The {PASCAL} {V}isual {O}bject {C}lasses {C}hallenge 2012 {(VOC2012)}
  {R}esults.
\newblock
  http://www.pascal-network.org/challenges/VOC/voc2012/workshop/index.html

\bibitem{TLinCVPR15}
Lin, T.Y., Maire, M., Belongie, S., Bourdev, L., Girchick, R., Hays, J.,
  Perona, P., Ramanan, D., Zitnick, C.L., Dollar, P.:
\newblock Microsoft {COCO}: Common objects in context.
\newblock In: CVPR. (2015)

\bibitem{jia2014caffe}
Jia, Y., Shelhamer, E., Donahue, J., Karayev, S., Long, J., Girshick, R.,
  Guadarrama, S., Darrell, T.:
\newblock Caffe: Convolutional architecture for fast feature embedding.
\newblock arXiv preprint arXiv:1408.5093 (2014)

\bibitem{AMishraICCV09}
Mishra, A., Aloimonos, Y., Fah, C.L.:
\newblock Active segmentation with fixation.
\newblock In: ICCV. (2009)

\bibitem{JUijlingsIJCV13}
Uijlings, J.R.R., van~de Sande, K.E.A., Gevers, T., Smeulders, A.W.M.:
\newblock Selective search for object recognition.
\newblock International Journal of Computer Vision \textbf{104}(2) (2013)
  154--171

\end{thebibliography}
\end{document}